\documentclass[11pt,theapa]{article}
\usepackage{fullpage,theapa, rawfonts, graphicx,subcaption,fancyvrb,amsmath,url}
\pdfoutput=1 

\begin{document}

\title{Unweighted Stochastic Local Search can be Effective for Random CSP Benchmarks}

\author{Christopher D. Rosin \\
        christopher.rosin@gmail.com \\
        Amara Health Analytics, San Diego, California, USA}
\date{September 2014}

\maketitle

\begin{abstract}

We present ULSA, a novel stochastic local search algorithm for random binary constraint satisfaction problems (CSP). 
ULSA is many times faster than the prior state of the art on a widely-studied suite of random CSP benchmarks.
Unlike the best previous methods for these benchmarks, ULSA is a simple unweighted method that does not require dynamic adaptation of weights or penalties.
ULSA obtains new record best solutions satisfying 99 of 100 variables in the challenging frb100-40 benchmark instance.

\end{abstract}

\section{Random Binary CSPs}
\label{introsection}

Randomly generated instances of constraint satisfaction problems (CSP) have
provided widely-used benchmarks in the development of solvers for CSP, Boolean satisfiability (SAT), and maximum independent set (MIS).
While some interesting instances of CSP and SAT eventually turn out to be reasonably easy for appropriate methods \cite{minconflicts,cnfsateasyinstances}, 
{\em reliably hard} instances of random CSP can be obtained from the ``phase transition'' boundary between
satisfiable and unsatisfiable regions of parameterized distributions of instances \cite{cspphasetransition}.
For appropriate distributions, most of these phase transition instances are exponentially hard for a
standard baseline approach of tree-like resolution \cite{rbresolution}.
And empirically these phase transition instances are challenging for all known methods.

A discrete {\em constraint satisfaction problem} (CSP) consists of $n$ variables $x_1 \ldots x_n$.
Each variable $x_i$ can be assigned a value from its finite domain $D_i$.  In this paper,
all variables in an instance have the same domain size.
An instance specifies a set of {\em constraints}, where each constraint identifies $k$ variables and restricts
them to an allowed subset of combinations of values from the cross product of
their domains.  Here, we deal with {\em binary} CSPs in which $k=2$, and we refer to the two variables
in a constraint as {\em endpoints} of the constraint.  If two endpoints hold values that violate
the constraint, we refer to it as a {\em conflict}.  A {\em solution} to the CSP is
a setting to the variables in which there are no conflicts.  We also refer to partial solutions as meeting a {\em target} of $T \leq n$
if there is a subset of $T$ variables such that no conflict has both endpoints in this subset.

Several models have been developed for generating hard random CSP instances \cite{modelabc}.  
{\em Model RB} addresses shortcomings of early models \cite{modelrb}, and has received wide attention.
To generate a random binary CSP with $n$ variables, Model RB draws (with replacement) $rn \ln(n)$ variable pairs to constrain.
For each of these variable pairs, the constraint disallows a fraction $p$ of pairs of values; the specific
pairs are drawn randomly (without replacement), independently for each constraint.  With $n$ variables, 
a phase transition occurs (asymptotically) when domain size is $n^{0.8}$, $r=0.8/(\ln(4)-\ln(3))$, and $p=0.25$ \cite{rbphasetransition}.  
It is also possible to modify Model RB slightly to generate {\em forced-satisfiable} instances that are
guaranteed to have a solution, by rejecting any constraints that are inconsistent with a preselected solution; 
such forced-satisfiable instances have very similar properties to the
original Model RB \cite{modelrbforced}.  

A particular benchmark suite of forced-satisfiable Model RB instances
\cite{bhoslib,bhoslibgraph}, denoted {\em frbX-Y} for X variables and domain size Y, has been used repeatedly in CSP solving competitions \cite{CSPcompetition2005,CSPcompetition2006,CSPcompetition2008,CSPcompetition2009}, SAT solving competitions \cite{SATcompetition}, and many published CSP and SAT studies \cite{earlyfrbusage,modelrbforced,caisat}.  
These benchmarks have been very challenging in CSP and SAT competitions; for example, instance frb59-26-2 appeared in at least 6 competitions from 2004-2011 without ever being solved by any of the competitors.

This benchmark suite has had particular importance in the development of MIS solvers \cite{ewls,cls,numvc}.\footnote{Instead of ``maximum independent set'' or ''MIS'', some studies refer to 
the equivalent problems of ``minimum vertex cover'' or ``maximum clique.''}
Given an undirected graph,
the MIS problem asks for the largest subset of vertices such that no edge has both
endpoints in the subset.  
For MIS, the graph formulation of this benchmark suite (known as {\em BHOSLIB}) relies
on a straightforward reduction from binary CSP to MIS: given a CSP instance, the corresponding MIS instance
has a vertex for each possible domain value of each variable, an edge for each pair
of values disallowed by a constraint, and an edge for each pair of values for the same variable
since a variable can only hold one value.\footnote{We confirmed (by inspection of subset of the BHOSLIB MIS instances that had previously been made available in CSP form) that the BHOSLIB MIS instance vertices are numbered sequentially starting with all of the domain values for one CSP variable, then all of the domain values for a second CSP variable, and so on.  This made it straightforward to recover the binary CSP instances from their MIS versions.  Had this straightforward recovery not been possible (e.g. due to random permutation of vertices in the MIS instance), we also confirmed that it is feasible to recover the CSP variables from the MIS instance with brief runs of a max clique heuristic (NuMVC operating on the inverse of the graph) to identify the large cliques in the MIS instance that represent the domain values of the CSP variables.}  Independent sets of size $n$ correspond to solutions to the original $n$-variable CSP instance, and
independent sets of size $T \leq n$ correspond to partial CSP solutions meeting a target of $T$.
As MIS solvers have become more capable via improved algorithms and increased computer speed,
most of the MIS benchmarks in the older DIMACS suite are readily solved, making BHOSLIB an important
source of newer challenging MIS benchmarks \cite{richter,cls,numvc}.  The importance
of BHOSLIB extends to use in the development of MIS solvers for practical biomedical applications \cite{PMC3564766,PMC3995366}.
Substantially improved MIS results on DIMACS and BHOSLIB instances have come through
use of stochastic local search algorithms that dynamically adapt weights or penalties on vertices \cite{dlsmc,cls}, edges \cite{ewls,numvc}, or both \cite{vertexedgeweighting}.
MIS solvers using such weighted search techniques have yielded the best results for the BHOSLIB benchmarks.

\section{NuMVC}

In comparisons below, we use the MIS algorithm NuMVC \cite{numvc} to represent the state of the art in 
solving the BHOSLIB benchmarks.  NuMVC and its predecessors by the same authors were
the first to achieve an independent set of 98 nodes on the large challenging frb100-40 benchmark \cite{ewls}.  
On moderate-sized BHOSLIB instances, NuMVC succeeds more reliably at speeds at least several times faster than earlier MIS methods by other authors \cite{numvc}.
Slightly better results on BHOSLIB (up to 20\% faster on some consistently-solved instances) have been reported
for a method that builds on the techniques used in NuMVC \cite{vertexedgeweighting}, but this does not materially change our conclusions below.

NuMVC is a {\em stochastic local search} algorithm.  Stochastic local search randomly selects a small localized change to a current candidate
solution, and then repeats this many times to traverse the search space \cite{hoosbook}.
The stochastic local search state in NuMVC is based on a current candidate subset of the vertices.  Each iteration, NuMVC selects a vertex to add
to the candidate subset; this may result in additional edges with both endpoints in the candidate subset.  
It then randomly selects such an edge and eliminates
it by removing one of its endpoints from the subset.  Each iteration, weights are increased on each edge with both endpoints in the candidate subset.
The choice of vertices to add and remove from the
subset is then biased to favor elimination of edges that have high weight.  Edge weights are maintained and adapted over
long periods consisting of many iterations.

\section{ULSA}

\begin{figure}
\centering
\small{
\begin{Verbatim}[frame=single,framesep=2mm,commandchars=\\\{\},codes={\catcode`$=3\catcode`^=7}]
Initialize n=0, and initialize t[v]=0 for each variable v
FOREACH variable v in randomly permuted order:
   Initialize x[v] to the value which minimizes conflicts (break ties randomly)

WHILE (there exists at least one conflict)
   Choose a random conflict; denote its endpoints (i,j) such that t[i]<=t[j]
   IF (x[i] could change without increasing conflicts) OR (t[j]==n)
      THEN S=\{i\} ELSE S=\{i,j\}
   Change x[k] to u, choosing k$\in$S \& u to minimize conflicts (break ties randomly)
   n=n+1
   t[k]=n
\end{Verbatim}
}
\caption{{\bf ULSA.}  x[v] is the current value of variable v.  All variables start out uninitialized, and variables cannot participate in conflicts until they have been initialized.  ``Change'' means that the variable must be set to a new value that differs from its current value.  n counts iterations and t[j] is the iteration at which variable j last changed, so (t[j]==n) checks if variable j was the most recent to change.}
\label{ulsafig}
\end{figure}

ULSA (Unweighted Local Search Algorithm, Figure~\ref{ulsafig}) is a novel stochastic local search algorithm that 
operates directly on a binary CSP (as opposed to operating on the graph formulation used by MIS algorithms).  
ULSA's search state consists of a candidate setting of the variables, and timestamps indicating when each variable was last updated.
Each iteration, ULSA randomly selects a current conflict 
and eliminates it by setting one of its endpoints to a different value; this is comparable to NuMVC's elimination of a randomly selected edge.
ULSA preferentially selects the conflict endpoint with the older timestamp; NuMVC
uses a similar mechanism for tiebreaking in case of equal weights.
Unlike NuMVC though, ULSA does not maintain weights or other comparably long-term state information.  

For a variable that is to be updated, ULSA selects a value that minimizes the number of conflicts.
This is similar to the well-known {\em min-conflicts} heuristic \cite{minconflicts}.  A min-conflicts hillclimbing
heuristic, however, is subject to getting trapped in local optima from which it cannot escape without temporarily
worsening the total number of conflicts \cite{minconflicts}.  In fact, some of the 
original development of dynamic weighting techniques like those used in NuMVC came in the {\em breakout}
method for escaping from local optima that could trap min-conflicts \cite{breakout}.  Here we
take an alternative unweighted approach, which empirically appears to be well-matched to
the challenging phase transition instances of Model RB.  First, ULSA requires that a variable 
change away from its current value -- even if that worsens the number of conflicts.
Second, if the initially selected endpoint would indeed force a worsening of the total number of conflicts,
the neighborhood is expanded to also allow candidate updates from the other endpoint of the originally
selected conflict.
Running on the BHOSLIB instances, this {\em neighborhood expansion} is activated in 45-65\% of iterations, and
approximately 25\% of iterations end up forced into worsening the number of unsatisfied constraints (despite neighborhood expansion).
Such worsening moves are essential for escaping local optima.

\section{Experiments}

We compare ULSA to NuMVC on the BHOSLIB benchmarks, and then
evaluate ULSA's ability to reach new record-setting solutions for the large challenging frb100-40 instance.

\subsection{Performance Comparison}

\begin{table}
\caption{Comparison of ULSA and NuMVC on BHOSLIB instances.  Binary CSP benchmark instances designated frbX-Y have X variables with domain size Y; the graph formulation used by NuMVC has X*Y nodes.  The ``Target'' for ULSA is as defined in Section~\ref{introsection} for ULSA, and for NuMVC is the equivalent required independent set size; for all instances except frb100-40 the indicated Target requires perfect solutions.  Each method is run 100 times on each instance, and 100\% of these runs terminated successfully.}
\label{mainresults}
\centering
\begin{tabular}{| r | c | r | r | r | r |}
\hline
 & & \multicolumn{2}{|c|}{NuMVC} & \multicolumn{2}{|c|}{ULSA} \\ 
Instance & Target & Iterations & Time (s)  & Iterations & Time (s) \\ \hline\hline
frb40-19-1 & 40 & 205831 &     0.16         & 25921        & 0.01   \\ 
frb40-19-2 & 40 & 3668573 &    2.74         & 473473       & 0.07   \\ 
frb40-19-3 & 40 & 873459 &     0.65         & 254746       & 0.04   \\ 
frb40-19-4 & 40 & 2602852 &    1.94         & 1210809      & 0.17   \\ 
frb40-19-5 & 40 & 7830157 &    5.84         & 3334878      & 0.45   \\ \hline
frb45-21-1 & 45 & 1993837 &    1.77         & 661451       & 0.10    \\ 
frb45-21-2 & 45 & 4377381 &    3.86         & 1817130      & 0.26   \\ 
frb45-21-3 & 45 & 9277665 &    8.16         & 3587173      & 0.50    \\ 
frb45-21-4 & 45 & 2718138 &    2.40          & 454525       & 0.07   \\ 
frb45-21-5 & 45 & 7712590 &    6.81         & 6581368      & 0.93   \\ \hline
frb50-23-1 & 50 & 21863585 &   22.67        & 5773150      & 0.83   \\ 
frb50-23-2 & 50 & 127828690 &  133.13       & 24751652     & 3.57   \\ 
frb50-23-3 & 50 & 348131684 &  362.37       & 209047771    & 30.06  \\ 
frb50-23-4 & 50 & 5572256 &    5.81         & 683894       & 0.11   \\ 
frb50-23-5 & 50 & 16190769 &   16.79        & 568476       & 0.09   \\ \hline
frb53-24-1 & 53 & 744855652 &  840.84       & 146679577    & 21.31  \\ 
frb53-24-2 & 53 & 120712887 &  136.34       & 103704282    & 14.93  \\ 
frb53-24-3 & 53 & 25652046 &   28.98        & 3016649      & 0.45   \\ 
frb53-24-4 & 53 & 152974941 &  173.00          & 24812439     & 3.63   \\ 
frb53-24-5 & 53 & 24413503 &   27.59        & 4122454      & 0.62   \\ \hline
frb56-25-1 & 56 & 319912278 &  387.75       & 198415701    & 29.33  \\ 
frb56-25-2 & 56 & 390181420 &  473.64       & 290399706    & 42.77  \\ 
frb56-25-3 & 56 & 63685811 &   77.27        & 10637855     & 1.56   \\ 
frb56-25-4 & 56 & 23537869 &   28.59        & 15215815     & 2.27   \\ 
frb56-25-5 & 56 & 13686349 &   16.56        & 1611838      & 0.25   \\ \hline
frb59-26-1 & 59 & 827709031 &  1079.55      & 187448484    & 27.76  \\ 
frb59-26-2 & 59 & 2309186751 & 3009.04      & 236516360    & 35.06  \\ 
frb59-26-3 & 59 & 389945144 &  508.30        & 62055067     & 9.21   \\ 
frb59-26-4 & 59 & 535848770 &  699.94       & 215800597    & 32.31  \\ 
frb59-26-5 & 59 & 30809115 &   40.12        & 2429727      & 0.38   \\ \hline
frb100-40 & 97 & 357808999 &   1022.87      & 17971765     & 3.20    \\ \hline
\end{tabular}
\end{table}

BHOSLIB consists of 5 instances for each number of variables in $\{30,35,40,45,50,53,56,59\}$, plus a single challenge instance frb100-40 with 100 variables.
We omit the instances with 30 or 35 variables, as they are readily solved by a wide variety of methods.
Instances with 40 to 59 variables are run to complete solution.
The large frb100-40 challenge instance is run to find partial solutions with a target of 97 (equivalent to an independent set of size 97 in the MIS formulation):
NuMVC automatically seeks such partial solutions with increasing targets, and ULSA is configured for this instance to check all states with at most 8 conflicts to find
if removing at most 3 variables (to yield a subset of 97 variables) can eliminate all conflicts.  The cap of 8 conflicts was set empirically; while there could exist partial solutions meeting a target of 97 that have more than 8 conflicts, we observed that most partial solutions meeting this target have fewer than 8 conflicts.   The target of 97 provides a challenge that is tractable enough for both NuMVC and ULSA to allow comparison over 100 runs.  

NuMVC and ULSA are compiled with identical optimization options.\footnote{Refer to http://chrisrosin.com/randomcsp for ULSA source code and compilation details.}  
NuMVC is run using code published by NuMVC's author \cite{numvccode}.
NuMVC parameters are set to originally published values for the BHOSLIB instances \cite{numvc}. 

NuMVC and ULSA are each run 100 times on each instance.  We compare them
in terms of average solution time, as is common when benchmarking stochastic local search algorithms \cite{hoosbook,numvc}.
For consistent runtime measurement, runs take place in a standard reference environment, in which a Linux server with 32GB of memory and 
a 3.4GHz i7-4770 CPU is fully dedicated to one run at a time of NuMVC or ULSA (both are single-threaded).  
NuMVC and ULSA are similar enough that in addition to runtime we also compare them in terms of the average number of iterations; each iteration performs a single node swap in NuMVC, and reassignment of a single variable in ULSA.

Results are shown in Table~\ref{mainresults}.  
All runs complete successfully; no run becomes stuck in a local optimum from which it cannot escape. 
NuMVC results are consistent with those reported previously \cite{numvc}, except that
no timeout is used here (all runs go to successful completion) and speeds are modestly faster due to newer hardware.

\begin{figure}
\centering
\includegraphics{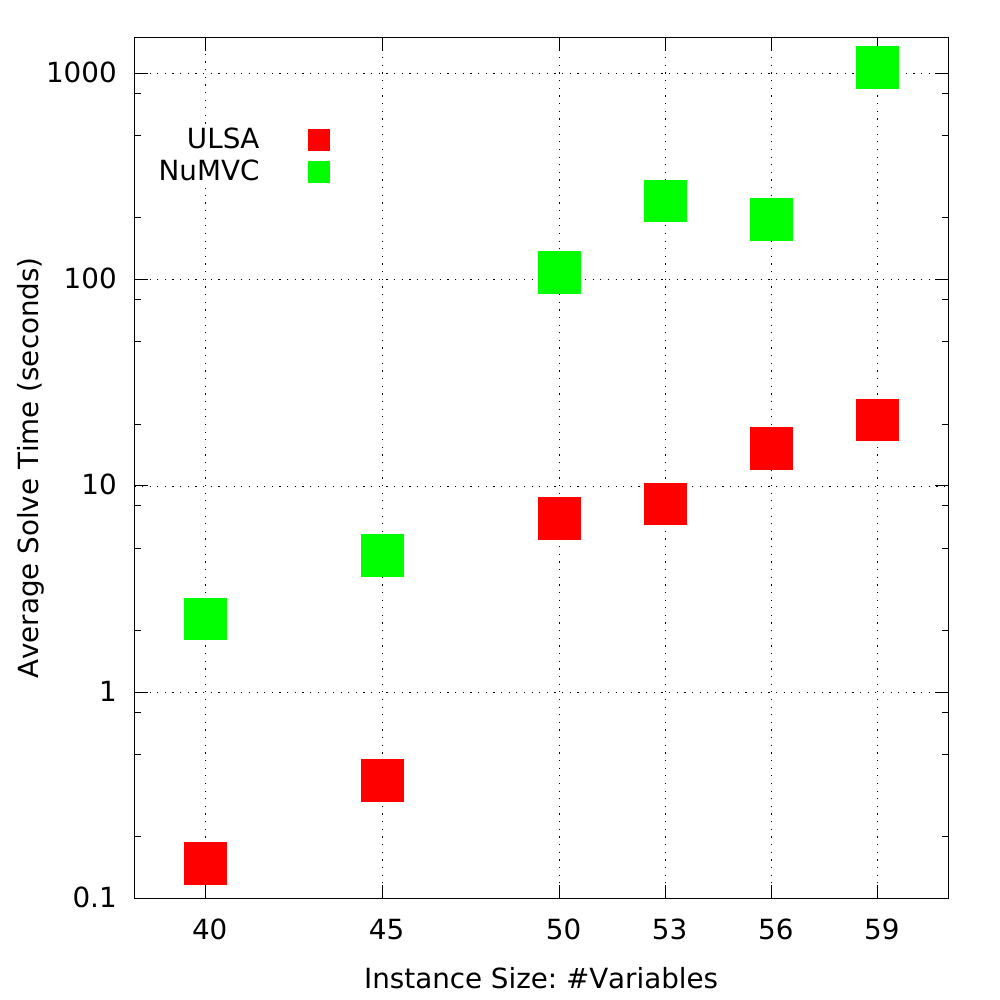}
\caption{ULSA and NuMVC: Comparison of Average Time to Solve}
\label{timescaling}
\end{figure}

ULSA solution times are at least 7x faster than NuMVC on every instance.
Figure~\ref{timescaling} shows solution times averaged over each group of five instances with the same number of variables.
ULSA's speed advantage over NuMVC on 40 variable instances is 15x, and on 59 variable instances is 51x.
On frb100-40 with partial solution target 97, ULSA's speed advantage is 320x.

Part of this speed advantage comes from efficient implementation enabled by the simple nature of ULSA.  
ULSA only considers values associated with 1-2 variables each iteration.
The inner loop of ULSA, which requires counting the number of conflicts resulting from changing the value x[k] to
candidate values u, is efficiently vectorized and computed simultaneously across many candidate values u.

While implementation speed gives an advantage to ULSA in time per iteration, ULSA also requires fewer iterations than NuMVC on every instance.
ULSA's advantage in average iteration count on 40 variable instances is 3x, and on 59 variable instances is 6x.
On the benchmark of partial solutions to frb100-40, ULSA's advantage in iteration count is 20x.

\subsection{Runtime Distributions}

\begin{figure}
\begin{subfigure}{0.3\textwidth}
\includegraphics{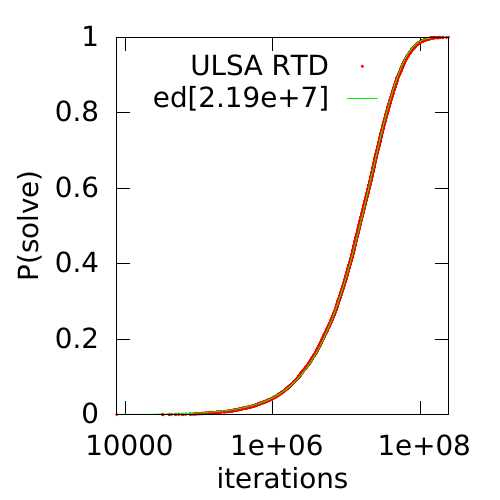}
\subcaption{}
\end{subfigure}
\begin{subfigure}{0.3\textwidth}
\includegraphics{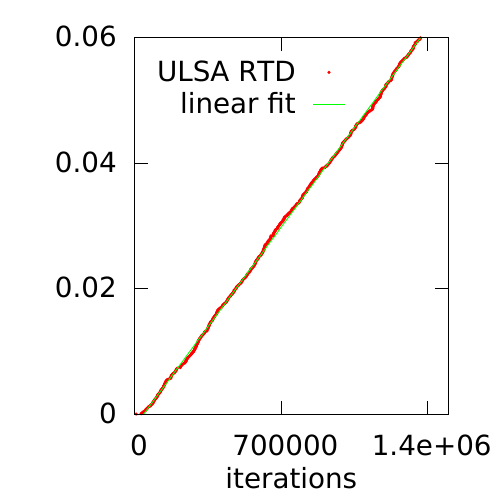}
\subcaption{}
\end{subfigure}
\begin{subfigure}{0.3\textwidth}
\includegraphics{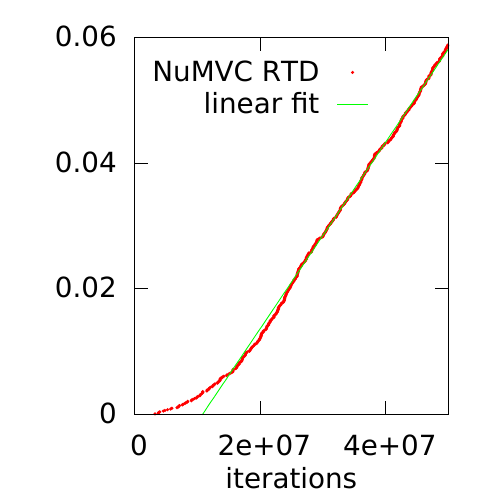}
\subcaption{}
\end{subfigure}
\caption{Runtime Distributions for ULSA and NuMVC for frb100-40 with target 97.
(a) Full ULSA Runtime Distribution (RTD), with exponential distribution (ed) fit of the form $1-e^{-x/m}$ (the fit and the data are nearly exactly overlapping and therefore difficult to separate visually).  This graph is shown on a logarithmic X axis.  (b) Early ULSA RTD, with linear fit shown to illustrate nearly constant solution probability per unit time.  This linear fit is consistent with the early part of (a)'s fit where the exponential distribution is approximately linear.  This graph is shown on a linear X axis.  (c) Early NuMVC RTD with fit to approximately-linear region, shown on a linear X axis.}
\label{stationaryfig}
\end{figure}

The runtime distribution (RTD) shows the observed cumulative probability of reaching a target solution, versus the number of iterations.
For stochastic local search algorithms, it is common practice to fit an exponential distribution to the RTD; such a distribution indicates that
the probability of solving in a given unit of time is reliably constant regardless of history, and implies
useful properties such as simple efficient parallelization \cite{hoosbook}.
The RTD for ULSA is well-fit by such an exponential distribution (Figure~\ref{stationaryfig}a).  NuMVC also has been observed to have this property \cite{numvc}.

Given the difference though between NuMVC's long-term adaptation of weights versus ULSA's lack of any comparably long-term state, we might expect to see this reflected in a difference in their RTDs.
Indeed, a closer look at the early part of the RTD shows that very quickly after initialization ULSA reaches a state of constant solution probability per unit time (Figure~\ref{stationaryfig}b), whereas 
NuMVC has a longer slow-progress initial period (Figure~\ref{stationaryfig}c), perhaps while NuMVC's edge weights are adapting to the instance.

\subsection{Best Results on frb100-40}

No reported method has ever solved the large challenge benchmark frb100-40.  ULSA has not solved frb100-40 either, but does improve upon the previous state of the art.

\subsubsection{Partial solutions satisfying 99 variables}

The best-reported previous results on frb100-40 reach a target of 98; an independent set of 98 nodes in the MIS instance \cite{ewls}.
We evaluate the ability of ULSA to improve upon this by running with a target of 99.
Candidate solutions with at most 5 conflicts are checked as to whether removing a single variable could remove all conflicts (which happens when that variable is an endpoint of all conflicts).  
Table~\ref{results99} shows results.\footnote{Unlike experiments above, these runs were not done in the reference environment.  Iterations per second were measured from the average of two separate runs in the reference environment, and then multiplied by the number of iterations reported here to obtain a derived solution time.}  All runs succeed in reaching the target of 99, providing new record-best solutions to this instance.  Single-threaded solution time averages 29.3 hours; running multiple threads on the multicore CPU used in the reference environment would easily yield an expected time of less than a day to obtain at least one 99.  The author of BHOSLIB stated ``I conjecture that in the next 20 years or more (from 2005) [frb100-40] can not be solved on a PC (or alike) in a reasonable time (e.g. 1 day)'' \cite{bhoslibgraph}.  This prediction stands, but it is at least possible (in 2014) to obtain in 1 day partial solutions that are one variable/vertex from optimal.

\begin{table}[h*]
\caption{ULSA results, obtaining new record-best solutions to frb100-40.}
\label{results99}
\centering
\begin{tabular}{| l | l | l | l | l | l |}
\hline
Instance & Target & Runs & Success & Average Iterations & Average Time \\ \hline\hline
frb100-40 & 99 & 16 & 100\% & $6.27\times10^{11}$ & 29.3 hours\\ \hline
\end{tabular}
\end{table}

\subsubsection{Partial solutions with few conflicts}

\begin{figure}
\centering
\includegraphics{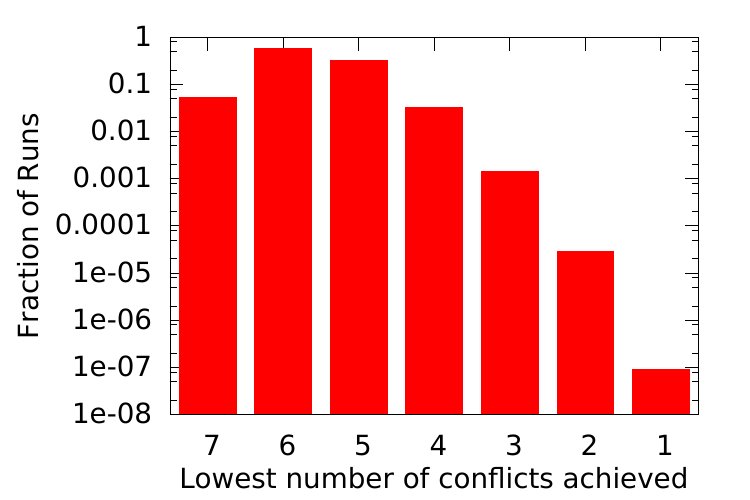}
\caption{Distribution of best results achieved in 3M iteration runs of ULSA on frb100-40}
\label{edgedistrib}
\end{figure}

We separately evaluate the ability of ULSA to obtain partial solutions that are close to optimal in the sense that they have few conflicts.
For this purpose, independent restarts of ULSA on frb100-40 are each run for 3 million iterations, recording the partial solution with the fewest conflicts during a run.
Figure~\ref{edgedistrib} shows the distribution.  Of 56 million runs, only five obtained a single-conflict partial solution.  Three of the five single-conflict solutions found by independent runs were equivalent.

\section{Limitations}

ULSA is not a general-purpose solver; it is designed for use with satisfiable random binary CSPs on the phase transition, particularly forced-satisfiable instances from Model RB.  
This is in contrast to other approaches like NuMVC which have been benchmarked on Model RB instances but are designed for and evaluated on a broader range of problems.

One potential failure mode for ULSA is getting stuck in local optima.  
Even for phase transition instances from Model RB, there is no guarantee this won't happen.  
It was never observed to occur in any experiments above though, and even if it did occasional restarts \cite{csprestarts} should mitigate it.
Outside of random CSPs near the phase transition that ULSA is designed for, however, this issue can limit performance.

\section{Conclusion}

The BHOSLIB benchmarks of Model RB are widely used as reliably challenging phase transition instances of random binary CSPs.
The most successful methods have been those like NuMVC that dynamically adapt weights on edges and/or vertices in graph formulations of these instances.
We have demonstrated ULSA, a simple parameterless unweighted stochastic local search algorithm for random binary CSPs.
Compared to a previous state of the art method on these benchmarks, ULSA is at least 7x faster on every instance and over 100x faster on
some instances, with improvements in both the
number of local search iterations and the time per iteration.
ULSA is the first method to reach a target of 99 on the challenging frb100-40 instance.

\section{Acknowledgments}
Thanks to Mark Land for helpful suggestions.

\vskip 0.2in
\bibliography{rosin-ulsa-2014}

\begin{thebibliography}{}

\bibitem[\protect\BCAY{Abraham\ \BBA\ Diaz}{Abraham\ \BBA\
  Diaz}{2014}]{PMC3995366}
Abraham, K.\BBACOMMA\  \BBA\ Diaz, C. \BBOP2014\BBCP.
\newblock \BBOQ Identifying large sets of unrelated individuals and unrelated
  markers\BBCQ\
\newblock {\Bem Source Code Biol Med}, {\Bem 9}, 6.

\bibitem[\protect\BCAY{Boussemart, Hemery,\ \BBA\ Lecoutre}{Boussemart
  et~al.}{2005}]{CSPcompetition2005}
Boussemart, F., Hemery, F., \BBA\ Lecoutre, C. \BBOP2005\BBCP\
\newblock In {\Bem CPAI-2005 Competition}, \BPGS\ 7--26.

\bibitem[\protect\BCAY{Bull, Muldoon,\ \BBA\ Doig}{Bull
  et~al.}{2013}]{PMC3564766}
Bull, S., Muldoon, M., \BBA\ Doig, A. \BBOP2013\BBCP.
\newblock \BBOQ Maximising the size of non-redundant protein datasets using
  graph theory\BBCQ\
\newblock {\Bem PLoS One}, {\Bem 8}, e55484.

\bibitem[\protect\BCAY{Cai}{Cai}{2011}]{numvccode}
Cai, S. \BBOP2011\BBCP.
\newblock \BBOQ {N}u{M}{V}{C} source\BBCQ\
\newblock Available from \url{http://lcs.ios.ac.cn/~caisw/MVC.html}.

\bibitem[\protect\BCAY{Cai, Su,\ \BBA\ Chen}{Cai et~al.}{2010}]{ewls}
Cai, S., Su, K., \BBA\ Chen, Q. \BBOP2010\BBCP.
\newblock \BBOQ {E}{W}{L}{S}: A new local search for minimum vertex cover\BBCQ\
\newblock In {\Bem AAAI-2010}, \BPGS\ 45--50.

\bibitem[\protect\BCAY{Cai, Su, Luo,\ \BBA\ Sattar}{Cai et~al.}{2013}]{numvc}
Cai, S., Su, K., Luo, C., \BBA\ Sattar, A. \BBOP2013\BBCP.
\newblock \BBOQ Nu{M}{V}{C}: An efficient local search algorithm for minimum
  vertex cover\BBCQ\
\newblock {\Bem JAIR}, {\Bem 46}, 687--716.

\bibitem[\protect\BCAY{Cai, Su,\ \BBA\ Sattar}{Cai et~al.}{2011}]{caisat}
Cai, S., Su, K., \BBA\ Sattar, A. \BBOP2011\BBCP.
\newblock \BBOQ A new local search strategy for {S}{A}{T}\BBCQ\
\newblock In {\Bem CSPSAT-11}.

\bibitem[\protect\BCAY{Fang, Chu, Qiao, Feng,\ \BBA\ Xu}{Fang
  et~al.}{2014}]{vertexedgeweighting}
Fang, S., Chu, Y., Qiao, K., Feng, X., \BBA\ Xu, K. \BBOP2014\BBCP.
\newblock \BBOQ Combining edge weight and vertex weight for minimum vertex
  cover problem\BBCQ\
\newblock In {\Bem FAW 2014}, \BPGS\ 71--81.

\bibitem[\protect\BCAY{Hoos\ \BBA\ St{\"u}tzle}{Hoos\ \BBA\
  St{\"u}tzle}{2004}]{hoosbook}
Hoos, H.\BBACOMMA\  \BBA\ St{\"u}tzle, T. \BBOP2004\BBCP.
\newblock {\Bem Stochastic Local Search}.
\newblock Morgan Kaufmann/Elsevier.

\bibitem[\protect\BCAY{Mehta\ \BBA\ van Dongen}{Mehta\ \BBA\ van
  Dongen}{2005}]{earlyfrbusage}
Mehta, D.\BBACOMMA\  \BBA\ van Dongen, M. \BBOP2005\BBCP.
\newblock \BBOQ Static value ordering heuristics for constraint satisfaction
  problems\BBCQ\
\newblock In {\Bem CPAI-2005}, \BPGS\ 65--78.

\bibitem[\protect\BCAY{Minton, Johnston, Philips,\ \BBA\ Laird}{Minton
  et~al.}{1990}]{minconflicts}
Minton, S., Johnston, M., Philips, A., \BBA\ Laird, P. \BBOP1990\BBCP.
\newblock \BBOQ Solving large-scale constraint satisfaction and scheduling
  using a heuristic repair method\BBCQ\
\newblock In {\Bem AAAI-90}, \BPGS\ 17--24.

\bibitem[\protect\BCAY{Morris}{Morris}{1993}]{breakout}
Morris, P. \BBOP1993\BBCP.
\newblock \BBOQ Breakout method for escaping local minima\BBCQ\
\newblock In {\Bem AAAI-93}, \BPGS\ 40--45.

\bibitem[\protect\BCAY{Prosser}{Prosser}{1996}]{cspphasetransition}
Prosser, P. \BBOP1996\BBCP.
\newblock \BBOQ An empirical study of phase transitions in binary constraint
  satisfaction problems\BBCQ\
\newblock {\Bem Artificial Intelligence}, {\Bem 81}, 81--109.

\bibitem[\protect\BCAY{Pullan\ \BBA\ Hoos}{Pullan\ \BBA\ Hoos}{2006}]{dlsmc}
Pullan, W.\BBACOMMA\  \BBA\ Hoos, H. \BBOP2006\BBCP.
\newblock \BBOQ Dynamic local search for {M}ax {C}lique\BBCQ\
\newblock {\Bem JAIR}, {\Bem 25}, 159--185.

\bibitem[\protect\BCAY{Pullan, Mascia,\ \BBA\ Brunato}{Pullan
  et~al.}{2011}]{cls}
Pullan, W., Mascia, F., \BBA\ Brunato, M. \BBOP2011\BBCP.
\newblock \BBOQ Cooperating local search for the maximum clique problems\BBCQ\
\newblock {\Bem J Heuristics}, {\Bem 17}, 181--199.

\bibitem[\protect\BCAY{Richter, Helmert,\ \BBA\ Gretton}{Richter
  et~al.}{2007}]{richter}
Richter, S., Helmert, M., \BBA\ Gretton, C. \BBOP2007\BBCP.
\newblock \BBOQ A stochastic local search approach to vertex cover\BBCQ\
\newblock In {\Bem KI-2007}, \BPGS\ 412--426.

\bibitem[\protect\BCAY{Smith\ \BBA\ Dyer}{Smith\ \BBA\ Dyer}{1996}]{modelabc}
Smith, B.\BBACOMMA\  \BBA\ Dyer, M. \BBOP1996\BBCP.
\newblock \BBOQ Locating the phase transition in binary constraint satisfaction
  problems\BBCQ\
\newblock {\Bem Artificial Intelligence}, {\Bem 81}, 155--181.

\bibitem[\protect\BCAY{van Dongen, Lecoutre,\ \BBA\ Roussel}{van Dongen
  et~al.}{2006}]{CSPcompetition2006}
van Dongen, M., Lecoutre, C., \BBA\ Roussel, O. \BBOP2006\BBCP.
\newblock \BBOQ {C}{S}{P} 2006 competition\BBCQ\
\newblock
  \url{http://www.cril.univ-artois.fr/CPAI06/round2/results/results.php?idev=6}.

\bibitem[\protect\BCAY{van Dongen, Lecoutre,\ \BBA\ Roussel}{van Dongen
  et~al.}{2008}]{CSPcompetition2008}
van Dongen, M., Lecoutre, C., \BBA\ Roussel, O. \BBOP2008\BBCP.
\newblock \BBOQ {C}{S}{P} 2008 competition\BBCQ\
\newblock
  \url{http://www.cril.univ-artois.fr/CPAI08/results/results.php?idev=15}.

\bibitem[\protect\BCAY{van Dongen, Lecoutre,\ \BBA\ Roussel}{van Dongen
  et~al.}{2009}]{CSPcompetition2009}
van Dongen, M., Lecoutre, C., \BBA\ Roussel, O. \BBOP2009\BBCP.
\newblock \BBOQ {C}{S}{P} 2009 competition\BBCQ\
\newblock
  \url{http://www.cril.univ-artois.fr/CSC09/results/results.php?idev=30}.

\bibitem[\protect\BCAY{van Maaren\ \BBA\ Franco}{van Maaren\ \BBA\
  Franco}{2014}]{SATcompetition}
van Maaren, H.\BBACOMMA\  \BBA\ Franco, J. \BBOP2014\BBCP.
\newblock \BBOQ The international {S}{A}{T} competitions\BBCQ\
\newblock \url{http://www.satcompetition.org}.

\bibitem[\protect\BCAY{Wallace}{Wallace}{1996}]{csprestarts}
Wallace, R. \BBOP1996\BBCP.
\newblock \BBOQ Analysis of heuristic methods for partial constraint
  satisfaction problems\BBCQ\
\newblock In {\Bem CP96}, \BPGS\ 482--496.

\bibitem[\protect\BCAY{Williams, Gomes,\ \BBA\ Selman}{Williams
  et~al.}{2003}]{cnfsateasyinstances}
Williams, R., Gomes, C., \BBA\ Selman, B. \BBOP2003\BBCP.
\newblock \BBOQ Backdoors to typical case complexity\BBCQ\
\newblock In {\Bem IJCAI-03}, \BPGS\ 1173--1178.

\bibitem[\protect\BCAY{Xu}{Xu}{2003}]{bhoslib}
Xu, K. \BBOP2003\BBCP.
\newblock \BBOQ Forced satisfiable {C}{S}{P} and {S}{A}{T} benchmarks of
  {M}odel {R}{B}\BBCQ\
\newblock \url{http://www.nlsde.buaa.edu.cn/~kexu/benchmarks/benchmarks.htm}.

\bibitem[\protect\BCAY{Xu}{Xu}{2005}]{bhoslibgraph}
Xu, K. \BBOP2005\BBCP.
\newblock \BBOQ {B}{H}{O}{S}{L}{I}{B}: Benchmarks with hidden optimum solutions
  for graph problems\BBCQ\
\newblock
  \url{http://www.nlsde.buaa.edu.cn/~kexu/benchmarks/graph-benchmarks.htm}.

\bibitem[\protect\BCAY{Xu, Boussemart, Hemery,\ \BBA\ Lecoutre}{Xu
  et~al.}{2007}]{modelrbforced}
Xu, K., Boussemart, F., Hemery, F., \BBA\ Lecoutre, C. \BBOP2007\BBCP.
\newblock \BBOQ Random constraint satisfaction: Easy generation of hard
  (satisfiable) instances\BBCQ\
\newblock {\Bem Artificial Intelligence}, {\Bem 171}, 514--534.

\bibitem[\protect\BCAY{Xu\ \BBA\ Li}{Xu\ \BBA\ Li}{2000}]{modelrb}
Xu, K.\BBACOMMA\  \BBA\ Li, W. \BBOP2000\BBCP.
\newblock \BBOQ Exact phase transitions in random constraint satisfaction
  problems\BBCQ\
\newblock {\Bem JAIR}, {\Bem 12}, 93--103.

\bibitem[\protect\BCAY{Xu\ \BBA\ Li}{Xu\ \BBA\ Li}{2003}]{rbphasetransition}
Xu, K.\BBACOMMA\  \BBA\ Li, W. \BBOP2003\BBCP.
\newblock \BBOQ Many hard examples in exact phase transitions with application
  to generating hard satisfiable instances\BBCQ\
\newblock {\Bem CoRR}, cs.CC/0302001.

\bibitem[\protect\BCAY{Xu\ \BBA\ Li}{Xu\ \BBA\ Li}{2006}]{rbresolution}
Xu, K.\BBACOMMA\  \BBA\ Li, W. \BBOP2006\BBCP.
\newblock \BBOQ Many hard examples in exact phase transitions\BBCQ\
\newblock {\Bem Theoretical Computer Science}, {\Bem 355}, 291--302.

\end{thebibliography}
\bibliographystyle{theapa}

\end{document}